\documentstyle[epsf]{kbsjrnl}                    % Computer Modern Fonts
\begin{document}

%%\draft % Optional, will cause a line at the bottom of each page
%% with the words `Draft' and the time and date that the article
%% was LaTeXed. Will also double space text.

%%%%% To be entered at Kluwers: =====>>

\journame{}
\volnumber{}
\issuenumber{}
\issuemonth{}
\volyear{}

%% Do not delete either of the following two commands.
%% Please supply facing curly brackets for the part you
%% are not using for this article.
\received{}\revised{}

\authorrunninghead{}
\titlerunninghead{}

%\setcounter{page}{275} %% Optional, uncomment to set page number.
%%%%  <<==== End of commands to be entered at Kluwers 

%----------------------------------------------------------------------
%%  Authors, start here ==>>

\renewcommand{\topfraction}{1.0}
\renewcommand{\bottomfraction}{1.0}
\renewcommand{\textfraction}{0.05} \renewcommand{\floatpagefraction}{0.10} 
\makeatletter
\input{myfullname.sty}
\makeatother

\newcommand{\ignore}[1]{}

\newcommand{\Winnow}{WinSpell}
\newcommand{\Bayes}{BaySpell}

\newcommand{\spelling}{context-sensitive spelling correction}
\newcommand{\Spelling}{Context-sensitive spelling correction}
\newcommand{\SPELLING}{Context-Sensitive Spelling Correction}

\newcommand{\ra}{\rightarrow}
\newcommand{\cmark}{\underline{\hspace*{1em}}}
\newcommand{\tag}[1]{\mbox{\sc #1}}
\newcommand{\weather}{\mbox{$\{{\it weather}, {\it whether\/}\}$}}
\newcommand{\hear}{\mbox{$\{{\it hear}, {\it here}\}$}}
\newcommand{\among}{\mbox{$\{{\it among}, {\it between\/}\}$}}
\newcommand{\their}{\mbox{$\{{\it their}, {\it there}, {\it they're\/}\}$}}
\newcommand{\desert}{\mbox{$\{{\it desert}, {\it dessert\/}\}$}}
\newcommand{\its}{\mbox{$\{{\it its}, {\it it's\/}\}$}}
\newcommand{\being}{\mbox{$\{{\it begin}, {\it being\/}\}$}}
\newcommand{\maybe}{\mbox{$\{{\it maybe}, \mbox{\it may be\/}\}$}}
\newcommand{\peace}{\mbox{$\{{\it peace}, {\it piece\/}\}$}}
\newcommand{\amount}{\mbox{$\{{\it amount}, {\it number\/}\}$}}
\newcommand{\desc}{\vphantom{y}} \newcommand{\stacklabel}[2]{\footnotesize\shortstack[l]{#1\desc \\ #2}}
\newcommand{\sig}[1]{\makebox[12pt][l]{\hspace*{3pt}$#1$}}
\newcommand{\comb}[2]{\mbox{$ \left(
  \begin{array}{@{}c@{}}\\[-16pt] #1 \\[-4pt] #2 \\[-16pt] \mbox{}
  \end{array} \right) $}}

\newdimen\digitwidth \setbox0=\hbox{\rm0} \digitwidth=\wd0
\newdimen\twodigits \setbox0=\hbox{\rm99.9} \twodigits=\wd0
\newcommand{\hundred}{\makebox[\twodigits]{100.0\hspace*{\digitwidth}}}

\newcommand{\cword}[1]{{\it #1\/}\hspace*{8pt} & \makebox[.01in]{within} &
  \hspace*{8pt}$\pm10$ words}
\newcommand{\colloc}[2]{#1 & \cmark & #2}
\newcommand{\feature}[1]{& \makebox[.01in]{#1} &}
\newcommand{\lhood}[3]{#1/#2$\,\rightarrow\,$#3}

\newcommand{\mnote}[1]
    {\marginpar        [{\tiny\begin{minipage}[t]{\marginparwidth}\raggedright#1                \end{minipage}}]        {\tiny\begin{minipage}[t]{\marginparwidth}\raggedright#1                \end{minipage}}}

\newcommand{\fd}[1]
   {${}\atop
         {{\left[
                \begin{tabular}{l@{$\;\;=\;\;$}l}
                       #1
                \end{tabular}
            \right]} 
           \atop {}}
     $}

\newcommand{\phrase}[2]
   {[$_{\rm #1}$#2]}

\newcounter{sentencectr}
\newcounter{sentencesubctr}

\renewcommand{\thesentencectr}{(\smainform{sentencectr})}
\renewcommand{\thesentencesubctr}{\thesentencectr\ssubform{sentencesubctr}}

\newcommand{\smainform}{\arabic}
\newcommand{\ssubform}{\roman}
\newcommand{\ssubpunc}{.{}}

\newcommand{\beginsentences}{ \begin{list}{(\thesentencectr)}
   {\usecounter{sentencesubctr}
    \setlength{\itemsep}{0 in}
    \addtolength{\leftmargin}{25 pt}
    \setlength{\parsep}{0 in}}} 
\def\endsentences{\end{list}}

\newcommand{\sitem}{\renewcommand{\thesentencesubctr}{(\smainform{sentencectr}}
                    \refstepcounter{sentencectr}
     \item[(\smainform{sentencectr})\hfill]}
\newcommand{\smainitem}{\renewcommand{\thesentencesubctr
                                    }{\thesentencectr\ssubform{sentencesubctr}}
                        \setcounter{sentencesubctr}{0}
                        \refstepcounter{sentencectr}
                        \refstepcounter{sentencesubctr}
     \item[\thesentencectr\hfill\ssubform{sentencesubctr}\ssubpunc]}
\newcommand{\ssubitem}{\refstepcounter{sentencesubctr}
     \item[\hfill\ssubform{sentencesubctr}\ssubpunc]}

\makeatletter            
\newcommand{\smainlabel}[1]{{\renewcommand{\@currentlabel}{\thesentencectr}\label{#1}}}

\newcommand{\ssublabel}[1]{{\renewcommand{\@currentlabel}{\ssubform{sentencesubctr}}\label{#1}}}
\makeatother

\newdimen\tabskip

\input{mypsboxit.sty}
\PScommands

\newdimen\maxbarlen \maxbarlen=2.0in
\newdimen\barlen
\makeatletter
\def\bar#1{\setbarlen{#1}\@ifnextchar[{\barcolor}{\barhollow}}
\def\barcolor[#1]{\raisebox{2.5pt}{\psboxit{box #1 setgray fill}  {\framebox[\barlen]{}}}}
\def\barhollow#1{\raisebox{2.5pt}{\framebox[\barlen]{}}}
\def\setbarlen#1{\barlen=0.01\maxbarlen \barlen=#1\barlen
  \advance\barlen by 2\fboxrule}
\makeatother

\newcommand{\barhollowrt}  {\raisebox{-1pt}{\epsfysize=7pt\epsfbox{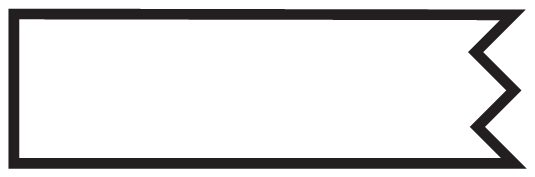}}}
\newcommand{\barlightrt}  {\raisebox{-1pt}{\epsfysize=7pt\epsfbox{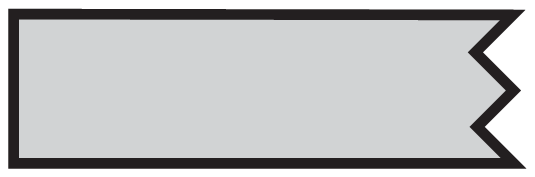}}}
\newcommand{\bardarkrt}  {\raisebox{-1pt}{\epsfysize=7pt\epsfbox{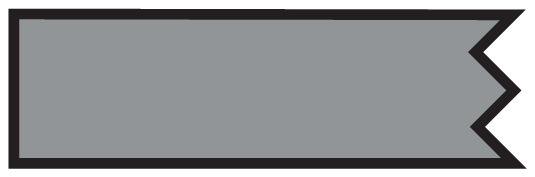}}}
\newcommand{\barhollowleft}  {\raisebox{-1pt}{\epsfysize=7pt\epsfbox{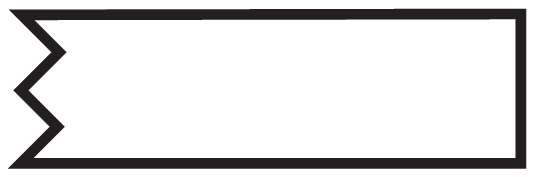}}}
\newcommand{\barlightleft}  {\raisebox{-1pt}{\epsfysize=7pt\epsfbox{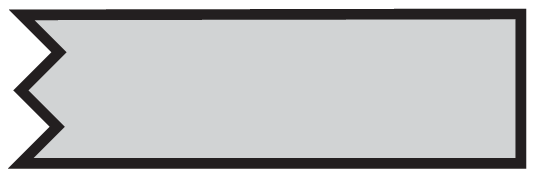}}}
\newcommand{\bardarkleft}  {\raisebox{-1pt}{\epsfysize=7pt\epsfbox{bar-dark-left.eps}}}

\newcommand{\bgcol}{@{}r@{\hspace*{0pt}}|@{\hspace*{0pt}}l@{}}
\newcommand{\bgheader}{\multicolumn{2}{c}{}}
\newdimen\bglabelwd \bglabelwd=\twodigits \advance\bglabelwd by 6pt
\newcommand{\bglab}[1]{\makebox[\bglabelwd]{#1}}

\begin{article}

\title{A Winnow-Based Approach to \SPELLING\thanks{An earlier version of
this work appeared in ICML'96.}}

\author{Andrew R. Golding}
\email{golding@merl.com}
\affil{MERL -- A Mitsubishi Electric Research Laboratory,
201 Broadway,
Cambridge, MA  02139}

\author{Dan Roth}
\email{danr@cs.uiuc.edu}
\affil{Department of Computer Science,
University of Illinois -- Urbana/Champaign,
1304 W. Springfield Avenue,
Urbana, IL  61801}

\editor{Raymond J. Mooney and Claire Cardie}

\abstract{
A large class of machine-learning problems in natural language
require the characterization of linguistic context.
Two characteristic properties of such problems are
that their feature space is of very high dimensionality,
and their target concepts depend on only a small subset
of the features in the space.
Under such conditions, multiplicative weight-update algorithms
such as Winnow have been shown to have exceptionally good
theoretical properties.
In the work reported here, we present an algorithm
combining variants of Winnow and weighted-majority voting,
and apply it to a problem in the aforementioned class: \spelling.
This is the task of fixing spelling errors that happen to result
in valid words, such as substituting {\it to\/} for {\it too},
{\it casual\/} for {\it causal}, and so on.
We evaluate our algorithm, \Winnow,
by comparing it against \Bayes, a statistics-based method
representing the state of the art for this task.
We find:
(1)~When run with a full (unpruned) set of features,
\Winnow\ achieves accuracies significantly higher than \Bayes\ was able to
achieve in either the pruned or unpruned condition;
(2)~When compared with other systems in the literature,
\Winnow\ exhibits the highest performance;
(3)~While several aspects of \Winnow's architecture
contribute to its superiority over \Bayes,
the primary factor is that it is able to learn a better linear separator
than \Bayes\ learns;
(4)~When run on a test set drawn from a different corpus
than the training set was drawn from,
\Winnow\ is better able than \Bayes\ to adapt,
using a strategy we will present that combines
supervised learning on the training set
with unsupervised learning on the (noisy) test set.
}

\keywords{Winnow, multiplicative weight-update algorithms, \spelling,
Bayesian classifiers}

\section{Introduction} \label{sec:intro}

A large class of machine-learning problems in natural language
require the characterization of linguistic context.
Such problems include lexical disambiguation tasks such as
part-of-speech tagging and word-sense disambiguation;
grammatical disambiguation tasks such as prepositional-phrase attachment;
and document-processing tasks such as text classification
(where the context is usually the whole document).
Such problems have two distinctive properties.
First, the richness of the linguistic structures that must be represented
results in extremely high-dimensional feature spaces for the problems.
Second, any given target concept
depends on only a small subset of the features,
leaving a huge balance of features that are irrelevant
to that particular concept.
In this paper, we present a learning algorithm and an architecture
with properties suitable for this class of problems.

The algorithm builds on recently introduced
theories of multiplicative weight-update algorithms.
It combines variants of Winnow \cite{Littlestone88}
and Weighted Majority \cite{LittlestoneWa94}.
Extensive analysis of these algorithms in the COLT literature
has shown them to have exceptionally good behavior
in the presence of irrelevant attributes,
noise, and even a target function changing in time
\cite{Littlestone88,LittlestoneWa94,HerbsterWa95}.
These properties make them particularly well-suited
to the class of problems studied here.

While the theoretical properties of the Winnow family of algorithms
are well known, it is only recently that people have started
to test the claimed abilities of the algorithms in applications.
We address the claims empirically by applying our Winnow-based algorithm
to a large-scale real-world task in the aforementioned class of problems:
\spelling.

\Spelling\ is the task of fixing spelling errors
that result in valid words, such as {\it I'd like a peace of cake},
where {\it peace\/} was typed when {\it piece\/} was intended.
These errors account for anywhere from 25\% to over 50\% of observed
spelling errors \cite{kukich-survey}; yet they go undetected by
conventional spell checkers, such as Unix {\it spell}, which only flag
words that are not found in a word list.
\Spelling\ involves learning to characterize the linguistic contexts
in which different words, such as {\it piece\/} and {\it peace}, tend to occur.
The problem is that there is a multitude of features one might use
to characterize these contexts:
features that test for the presence of a particular word
nearby the target word;
features that test the pattern of parts of speech
around the target word; and so on.
For the tasks we will consider, the number of features
ranges from a few hundred to over 10,000.\footnote{
  We have tested successfully with up to 40,000
  features, but the results reported here use up to 11,000.}
While the feature space is large, however,
target concepts, such as ``a context in which {\it piece\/} can occur'',
depend on only a small subset of the features,
the vast majority being irrelevant to the concept at hand.
\Spelling\ therefore fits the characterization presented above,
and provides an excellent testbed for studying
the performance of multiplicative weight-update algorithms
on a real-world task.

To evaluate the proposed Winnow-based algorithm, \Winnow,
we compare it against \Bayes\ \cite{Golding95},
a statistics-based method that is among the most successful
tried for the problem.  We first compare \Winnow\ and \Bayes\ using
the heavily-pruned feature set that \Bayes\ normally uses (typically
10--1000 features).  \Winnow\ is found to perform comparably to
\Bayes\ under this condition.  When the full, unpruned feature set is
used, however, \Winnow\ comes into its own, achieving substantially
higher accuracy than it achieved in the pruned condition, and better
accuracy than \Bayes\ achieved in either condition.

To calibrate the observed performance of \Bayes\ and \Winnow,
we compare them to other methods reported in the literature.
\Winnow\ is found to significantly outperform all the other methods tried
when using a comparable feature set.

At their core, \Winnow\ and \Bayes\ are both linear separators.
Given this fundamental similarity between the algorithms,
we ran a series of experiments to understand why
\Winnow\ was nonetheless able to outperform \Bayes.
While several aspects of the \Winnow\ architecture
were found to contribute to its superiority,
the principal factor was that \Winnow\ simply learned
a better linear separator than \Bayes\ did.
We attribute this to the fact that the Bayesian linear separator
was based on idealized assumptions about the domain,
while Winnow was able to adapt, via its mistake-driven update rule,
to whatever conditions held in practice.

We then address the issue of dealing with a test set that is
dissimilar to the training set.  This arises in
\spelling, as well as related disambiguation tasks,
because patterns of word usage can vary widely across documents;
thus the training and test documents may be quite different.
After first confirming experimentally that performance
does indeed degrade for unfamiliar test sets,
we present a strategy for dealing with this situation.
The strategy, called {\it sup/unsup}, combines supervised learning on the
training set with unsupervised learning on the (noisy) test set.
We find that, using this strategy, both \Bayes\ and \Winnow\ are able
to improve their performance on an unfamiliar test set.
\Winnow, however, is found to do particularly well,
achieving comparable performance when using the strategy on an unfamiliar
test set as it had achieved on a familiar test set.

The rest of the paper is organized as follows:
the next section describes the task of \spelling.
We then present the Bayesian method that has been used for it.
The Winnow-based approach to the problem is introduced.
The experiments on \Winnow\ and \Bayes\ are presented.
The final section concludes.

\section{\Spelling}

With the widespread availability of spell checkers
to fix errors that result in non-words, such as {\it teh},
the predominant type of spelling error has become the kind
that results in a real, but unintended word;
for example, typing {\it there\/} when {\it their\/} was intended.
Fixing this kind of error requires a completely
different technology from that used in conventional spell checkers:
it requires analyzing the context to infer
when some other word was more likely to have been intended.
We call this the task of {\it \spelling}.
The task includes fixing not only ``classic'' types of spelling mistakes,
such as homophone errors (e.g., {\it peace\/} and {\it piece\/})
and typographic errors (e.g., {\it form\/} and {\it from\/}),
but also mistakes that are more commonly regarded
as grammatical errors (e.g., {\it among\/} and {\it between\/}),
and errors that cross word boundaries
(e.g., {\it maybe\/} and {\it may be\/}).

The problem has started receiving attention in the literature
only within about the last half-dozen years.
A number of methods have been proposed, either for
\spelling\ directly, or for related lexical
disambiguation tasks.  The methods include
word trigrams \cite{mdm91},
Bayesian classifiers \cite{gale-word-sense},
decision lists \cite{Yarowsky94},
Bayesian hybrids \cite{Golding95},
a combination of part-of-speech trigrams and Bayesian hybrids \cite{golsch96},
and, more recently,
transformation-based learning \cite{manbri97},
latent semantic analysis \cite{jonmar97},
and differential grammars \cite{powers97}.
While these research systems have gradually been achieving
higher levels of accuracy, we believe that
a Winnow-based approach is particularly promising for this problem,
due to the problem's need for a very large number of features
to characterize the context in which a word occurs,
and Winnow's theoretically-demonstrated ability to handle
such large numbers of features.

\subsection{Problem formulation} \label{sec:problem}

We will cast \spelling\ as a word disambiguation task.
The ambiguity among words is modelled by {\it confusion sets}.
A confusion set $C = \{W_1, \ldots, W_n\}$ means
that each word $W_i$ in the set is ambiguous with each other word.
Thus if $C = $\hear,
then when we see an occurrence of either {\it hear\/}
or {\it here\/} in the target document, we take it to
be ambiguous between {\it hear\/} and {\it here\/};
the task is to decide from the context which one was actually intended.
Acquiring confusion sets is an interesting problem in its own right;
in the work reported here, however, we take our confusion sets
largely from the list of ``Words Commonly Confused''
in the back of the Random House dictionary \cite{random-house},
which includes mainly homophone errors.
A few confusion sets not in Random House were added,
representing grammatical and typographic errors.

The Bayesian and Winnow-based methods for
\spelling\ will be described below in terms of their operation
on a single confusion set; that is, we will say how
they disambiguate occurrences of words $W_1$ through $W_n$.
The methods handle multiple confusion sets
by applying the same technique to each confusion set independently.

\subsection{Representation} \label{sec:rep}

A target problem in \spelling\ consists of
(i)~a sentence, and (ii)~a target word in that sentence to correct.
Both the Bayesian and Winnow-based algorithms studied here
represent the problem as a list of active features;
each active feature indicates the presence of a particular linguistic
pattern in the context of the target word.
We use two types of features:
{\it context words\/} and {\it collocations}.
Context-word features test for the presence of a particular
word within $\pm k$ words of the target word;
collocations test for a pattern of up to $\ell$ contiguous
words and/or part-of-speech tags\footnote{
  Each word in the sentence is tagged with its {\it set\/}
  of possible part-of-speech tags, obtained from a dictionary.
  For a tag to match a word, the tag must be a member
  of the word's tag set.}
around the target word.
In the experiments reported here, $k$ was set to~10
and $\ell$ to~2.
Examples of useful features for the confusion set \weather\ include:
\beginsentences
\sitem {\it cloudy\/} within $\pm10$ words     \label{sent:cloudy}
\sitem \cmark~{\it to\/} \tag{verb}            \label{sent:toverb}
\endsentences
Feature~\ref{sent:cloudy} is a context-word feature that tends to imply
{\it weather}.
Feature~\ref{sent:toverb} is a collocation that checks for
the pattern ``{\it to\/} \tag{verb}'' immediately after the target word,
and tends to imply {\it whether\/} (as in {\it I don't know
whether to laugh or cry\/}).

The intuition for using these two types of features
is that they capture two important, but complementary aspects of context.
Context words tell us what kind of words tend to appear
in the vicinity of the target word --- the ``lexical atmosphere''.
They therefore capture aspects of the context
with a wide-scope, semantic flavor,
such as discourse topic and tense.
Collocations, in contrast, capture the local syntax
around the target word.
Similar combinations of features have been used in related tasks,
such as accent restoration \cite{Yarowsky94}
and word sense disambiguation \cite{nglee96}.

We use a {\it feature extractor\/} to convert
from the initial text representation of a sentence
to a list of active features.  The feature extractor
has a preprocessing phase in which it
learns a set of features for the task.
Thereafter, it can convert a sentence into a list
of active features simply by matching its set of learned features
against the sentence.

In the preprocessing phase, the feature extractor learns a set of features
that characterize the contexts in which each word $W_i$
in the confusion set tends to occur.
This involves going through the training corpus,
and, each time a word in the confusion set occurs,
generating all possible features for the context ---
namely, one context-word feature for every distinct word within $\pm k$ words,
and one collocation for every way of expressing
a pattern of up to $\ell$ contiguous elements.
This gives an exhaustive list of all features found in the training set.
Statistics of occurrence of the features are collected in the process as well.

At this point, pruning criteria may be applied
to eliminate unreliable or uninformative features.
We use two criteria (which make use of the aforementioned
statistics of occurrence):
(1)~the feature occurred in practically none or all
of the training instances (specifically, it had fewer
than 10 occurrences or fewer than 10 non-occurrences); or
(2)~the presence of the feature is not
significantly correlated with the identity of the target word
(determined by a chi-square test at the 0.05 significance level).

\section{Bayesian approach} \label{sec:bayes}

Of the various approaches that have been tried for \spelling,
the Bayesian hybrid method, which we call \Bayes, has been among the
most successful, and is thus
the method we adopt here as the benchmark for comparison with \Winnow.
\Bayes\ has been described elsewhere \cite{Golding95}, and so will
only be briefly reviewed here; however, the version here uses an
improved smoothing technique, which is described below.

Given a sentence with a target word to correct,
\Bayes\ starts by invoking the feature extractor (Section~\ref{sec:rep})
to convert the sentence into a set $\cal F$ of active features.
\Bayes\ normally runs the feature extractor with pruning enabled.
To a first approximation, \Bayes\ then acts as a naive Bayesian classifier.
Suppose for a moment that we really were applying Naive Bayes.
We would then calculate the probability that each word $W_i$
in the confusion set is the correct identity of the target word,
given that features $\cal F$ have been observed,
by using Bayes' rule with the conditional independence assumption:
\[
P(W_i|{\cal F}) = \left( \prod_{\;f \in {\cal F}} P(f|W_i) \right)
    \frac{P(W_i)}{P({\cal F})}
\]
\noindent
where each probability on the right-hand side
is calculated by a maximum-likelihood estimate\footnote{
  The maximum-likelihood estimate of $P(f|W_i)$ is
  the number of occurrences of $f$ in the presence of $W_i$
  divided by the number of occurrences of $W_i$.}
(MLE) over the training set.
We would then pick as our answer
the $W_i$ with the highest $P(W_i|{\cal F})$.

\Bayes\ differs from the naive approach in two respects:
first, it does not assume conditional independence among features,
but rather has heuristics for detecting strong dependencies,
and resolving them by deleting features
until it is left with a reduced set $\cal F'$ of
(relatively) independent features, which are then used in place of
$\cal F$ in the equation above.
This procedure is called {\it dependency resolution}.

Second, to estimate the $P(f|W_i)$ terms,
\Bayes\ does not use the simple MLE,
as this tends to give likelihoods of 0.0 for rare features
(which are abundant in the task at hand),
thus yielding a useless answer of 0.0 for the posterior probability.
Instead, \Bayes\ performs smoothing
by interpolating between the MLE of $P(f|W_i)$
and the MLE of the unigram probability, $P(f)$.
Some means of incorporating a lower-order model in this way
is generally regarded as essential for good performance \cite{chegoo96}.
We use:
\[
P_{\rm interp}(f|W_i) =
  (1 - \lambda) P_{\rm ML}(f|W_i) + \lambda P_{\rm ML}(f)
\]
\noindent We set $\lambda$ to the probability that
the presence of feature $f$ is independent of the presence of word $W_i$;
to the extent that this independence holds,
$P(f)$ is an accurate (but more robust) estimate of $P(f|W_i)$.
We calculate $\lambda$ as the chi-square probability
that the observed association between $f$ and $W_i$ is due to chance.

The enhancement of smoothing,
and to a minor extent, dependency resolution,
greatly improve the performance
of \Bayes\ over the naive Bayesian approach.
(The effect of these enhancements
can be seen empirically in Section~\ref{sec:ablation}.)

\ignore{
Figure~\ref{fig:bayesexample} gives an example
of \Bayes\ (and the feature extractor)
for correcting \peace\ in the sentence
{\it John had a peace of cake}.

\begin{figure}
\begingroup\catcode`~=\active\def~{\kern\digitwidth}\unitlength=.001in
\begin{picture}(5000,4600)(0,1400)

\put(2500,5800){\makebox(0,0){John had a {\it peace\/} of cake.}}

\put(2500,5410){\makebox(0,0){Feature extraction}}
\put(1500,5300){\framebox(2000,200){}}
\put(2500,5650){\vector(0,-1){150}}
\put(2500,5300){\vector(0,-1){1450}}

\put(2500,4650){\makebox(0,0)[l]{
  $\left.\begin{tabular}{@{}r@{\ }c@{\ }l@{}}
  \colloc{}{{\it of\/} \tag{noun}$_{\rm sing}$} \\
  \colloc{}{{\it of\/} \tag{verb}} \\
  \colloc{\tag{det}}{\tag{prep}} \\[2pt]
  \cword{had} \\
  \cword{of} \\
  \multicolumn{3}{c}{$\vdots$}
  \end{tabular}\hspace*{.1in}\right\}$
}}

\put(4350,4650){\makebox(0,0){
  \small
  \begin{tabular}{@{}c@{}}
  20 \\ active \\ features
  \end{tabular}
}}
  
\put(2500,3750){\makebox(0,0){Dependency resolution}}
\put(1500,3650){\framebox(2000,200){}}
\put(2500,3650){\vector(0,-1){350}}
\put(2550,3500){\makebox(0,0)[l]{\small 4 (relatively) independent features}}

\newcommand{\adj}{\hspace*{-10pt}}
\put(2500,2600){\makebox(0,0){
  \begin{tabular}{@{}r@{\ }c@{\ }l @{\ \ \ } c @{\ \ \ } c @{} c @{} c@{}}
  \feature{\small Feature $f$} &
     {\small $P(f|{\it peace\/})$} & {\small $P(f|{\it piece\/})$} &
     {\small $P({\it peace\/}|\cal F)$} & {\small $P({\it piece\/}|\cal F)$} \\
  \hline
  \feature{Priors}
     & & & .517~~~~\adj & .483~~~~\adj \\[2pt]
  \colloc{}{{\it of\/} \tag{noun}$_{\rm sing}$}
     & \lhood{~1}{105}{.0095} & \lhood{47}{98}{.4796}
     & .00493~~\adj & .232~~~~\adj \\[2pt]
  \colloc{{\it a\/}}{} 
     & \lhood{~6}{105}{.0578} & \lhood{28}{98}{.2836}
     & .000285~\adj & .0657~~~\adj \\[2pt]
  \cword{had} 
     & \lhood{~9}{105}{.0766} & \lhood{~2}{98}{.0264}
     & .0000218\adj & .00173~~\adj \\[2pt]
  \cword{.}
     & \lhood{47}{105}{.4599} & \lhood{47}{98}{.4666}
     & .0000100\adj & .000808~\adj
  \end{tabular}
}}

\put(100,1900){\framebox(4800,1400)}
\put(150,3400){\makebox(0,0)[lb]{Bayesian calculation}}

\put(2500,1525){\makebox(0,0)[t]{Suggest {\it piece\/}}}
\put(2500,1900){\vector(0,-1){300}}

\put(0,1750){\dashbox{40}(5000,2250){}}
\put(50,4100){\makebox(0,0)[lb]{\Bayes}}

\end{picture}
\caption{\Bayes\ example.  The target problem is to correct {\it peace\/}
in the sentence {\it John had a peace of cake.}
The feature extractor converts the problem
into a list of active features.
\Bayes\ then performs dependency resolution,
reducing the 20 matching features to 4~(relatively) independent ones.
The last box shows the application of Bayes' rule
using these 4~features, together with the priors,
to calculate a posterior probability for each word in the confusion set.
Likelihoods are shown as MLE ratios (i.e., the number of word-feature
co-occurrences divided by the number of word occurrences),
along with the resulting smoothed values.
The word {\it piece\/} is found to have the greatest posterior probability,
0.000808, and is thus suggested by \Bayes.
}
\label{fig:bayesexample}
\endgroup
\end{figure}
}

\section{Winnow-based approach} \label{sec:winnow}

There are various ways to use a learning algorithm,
such as Winnow \cite{Littlestone88}, to do the task of \spelling.
A straightforward approach would be to learn, for each confusion set,
a discriminator that distinguishes specifically among the words in that set.
The drawback of this approach, however, is that the learning is then
applicable only to one particular discrimination task.
We pursue an alternative approach:
that of learning the contextual characteristics
of each word $W_i$ individually.
This learning can then be used to distinguish word $W_i$
from any other word, as well as to perform
a broad spectrum of other natural language tasks
\cite{Roth98}.
In the following, we briefly present the general approach,
and then concentrate on the task at hand, \spelling.

The approach developed is influenced by the Neuroidal system suggested
by \namecite{Valiant94}.
The system consists of a very large number of items,
in the range of $10^5$.
These correspond to high-level concepts, for which humans have words,
as well as lower-level predicates from which
the high-level ones are composed.
The lower-level predicates encode aspects of the current state
of the world, and are input to the architecture from the outside.
The high-level concepts are learned as functions of the lower-level predicates;
in particular, each high-level concept is learned by
a {\it cloud\/} or ensemble of classifiers.
All classifiers within the cloud learn the cloud's high-level concept
autonomously, as a function of the same lower-level predicates,
but with different values of the learning parameters.
The outputs of the classifiers are combined into an output
for the cloud using a variant
of the Weighted Majority algorithm \cite{LittlestoneWa94}.
Within each classifier,
a variant of the Winnow algorithm \cite{Littlestone88} is used.
Training occurs whenever the architecture interacts with the world,
for example, by reading a sentence of text;
the architecture thereby receives new values for its
lower-level predicates, which in turn serve as an example
for training the high-level ensembles of classifiers.
Learning is thus an on-line process that is done
on a continuous basis\footnote{
  For the purpose of the experimental studies presented here,
    we do not update the knowledge representation while testing.
    This is done to provide a fair comparison
  with the Bayesian method which is a batch approach.}
\cite{Valiant95}.

\begin{figure}
\vspace*{0in}  \begin{center}
\leavevmode\epsfbox{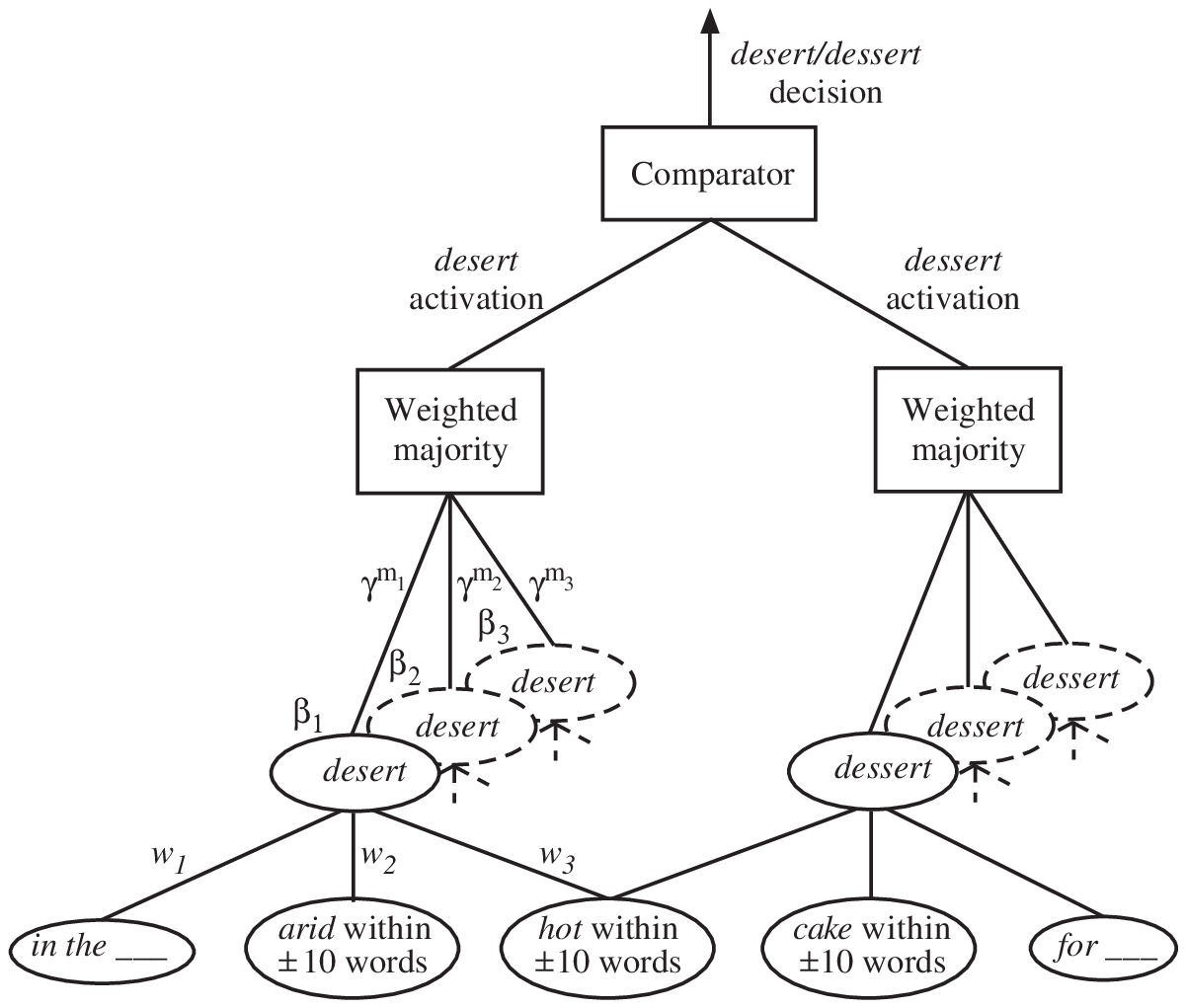}
\end{center}
\vspace*{0in}  \caption{Example of \Winnow\ network for \desert.
The five nodes in the bottom tier of the network correspond to features.
The two higher-level {\it clouds\/} of nodes
(each shown as overlapping bubbles suspended from a box)
correspond to the words in the confusion set.
The nodes within a cloud each run the Winnow algorithm in parallel
with a different setting of the demotion parameter, $\beta$,
and with their own copy of the input arcs and the weights on those arcs.
The overall activation level for each word in the confusion set
is obtained by applying a weighted majority algorithm
to the nodes in the word's cloud.
The word with the highest activation level is selected.
}
\label{fig:winnowarch}
\end{figure}

Figure~\ref{fig:winnowarch} shows the instantiation of the architecture
for \spelling, and in particular
for correcting the words \desert.
The bottom tier of the network consists of nodes for lower-level predicates,
which in this application correspond to features of the domain.
For clarity, only five nodes are shown;
thousands typically occur in practice.
High-level concepts in this application correspond to
words in the confusion set, here {\it desert\/} and {\it dessert}.
Each high-level concept appears as a {\it cloud\/} of nodes,
shown as a set of overlapping bubbles suspended from a box.
The output of the clouds is an activation level for each
word in the confusion set; a {\it comparator\/} selects
the word with the highest activation as the final result
for \spelling.

The sections below elaborate on the use of Winnow
and Weighted Majority in \Winnow,
followed by a discussion of the properties
of the architecture.

\subsection{Winnow} \label{sec:winnowalg}

The job of each classifier within a cloud of \Winnow\ is to
decide whether a particular word $W_i$ in the confusion set
belongs in the target sentence.
Each classifier runs the Winnow algorithm.
It takes as input a representation
of the target sentence as a set of active features,
and returns a binary decision as to whether
its word $W_i$ belongs in the sentence.
Let $\cal F$ be the set of active features;
and for each feature $f \in {\cal F}$,
let $w_f$ be the weight on the arc
connecting $f$ to the classifier at hand.
The Winnow algorithm then returns a classification of 1 (positive) iff:
$$ \sum_{f \in {\cal F}} w_f > \theta, $$
where $\theta$ is a threshold parameter.
In the experiments reported here, $\theta$ was set to 1.

Initially, the classifier has no connection
to any feature in the network.  Through training, however,
it establishes appropriate connections,
and learns weights for these connections.
A training example consists of a sentence,
represented as a set of active features,
together with the word $W_c$ in the confusion set
that is correct for that sentence.
The example is treated as a positive example
for the classifiers for $W_c$, and as a negative example
for the classifiers for the other words in the confusion set.

Training proceeds in an on-line fashion: an example is presented to
the system, the representation of the classifiers is updated, and the
example is then discarded.
The first step of training a classifier on an example
is to establish appropriate connections
between the classifier and the active features ${\cal F}$ of the example.
If an active feature $f \in {\cal F}$ is not already
connected to the classifier, and the sentence
is a {\it positive\/} example for the classifier
(that is, the classifier corresponds to the target word $W_c$ that 
occurs in the sentence),
then we add a connection between the feature and the classifier,
with a default weight of 0.1.
This policy of building connections on an as-needed basis
results in a {\it sparse\/} network with only those connections
that have been demonstrated to occur in real examples.
Note that we do not build any new connections
if the sentence is a {\it negative\/} example
for the classifier\footnote{
  This does not interfere with the subsequent updating
  of the weights --- conceptually, we treat a ``non-connection''
  as a link with weight 0.0, which will remain 0.0 after
  a multiplicative update.};
one consequence is that different words in a confusion set may have links
to different subsets of the possible features,
as seen in Figure~\ref{fig:winnowarch}.

The second step of training is to update the weights on the connections.
This is done using the Winnow update rule, which updates
the weights only when a mistake is made.
If the classifier predicts 0 for a positive example
(i.e., where 1 is the correct classification),
then the weights are promoted:
$$\forall f \in {\cal F},\, w_f \leftarrow \alpha \cdot w_f, $$
where $\alpha > 1$ is a promotion parameter.
If the classifier predicts 1 for a negative example
(i.e., where 0 is the correct classification),
then the weights are demoted:
$$\forall f \in {\cal F},\, w_f \leftarrow \beta \cdot w_f, $$
where $0 < \beta < 1$ is a demotion parameter.
In the experiments reported here,
$\alpha$ was set to 1.5, and $\beta$ was varied from 0.5 to 0.9
(see also Section~\ref{sec:weighted}).
In this way, weights on non-active features remain unchanged, and the
update time of the algorithm depends on the number of {\it active\/}
features in the current example, and not the total number of features
in the domain.  The use of a sparse architecture, as described above,
coupled with the representation of each example as a list of
{\it active\/} features
is reminiscent of the infinite attribute models of Winnow \cite{Blum92}.

\subsection{Weighted Majority} \label{sec:weighted}

Rather than evaluating the evidence for a given word $W_i$
using a single classifier, \Winnow\ combines evidence
from multiple classifiers; the motivation for doing so is discussed below.
Weighted Majority \cite{LittlestoneWa94}
is used to do the combination.
The basic approach is to run several classifiers in parallel
within each cloud to try to predict whether $W_i$
belongs in the sentence.  Each classifier uses different values of
the learning parameters, and therefore makes slightly different predictions.
The performance of each classifier is monitored,
and a weight is derived reflecting its observed prediction accuracy.
The final activation level output by the cloud
is a sum of the predictions of its member classifiers,
weighted by the abovementioned weights.

More specifically, we used clouds composed of five classifiers,
differing only in their values for the Winnow demotion
parameter $\beta$; values of 0.5, 0.6, 0.7, 0.8, and 0.9 were used.
The weighting scheme assigns to the $j$th classifier a weight
$\gamma^{m_j}$, where $0 < \gamma < 1$ is a constant,
and $m_j$ is the total number of mistakes made by the classifier so far.
The essential property is that the weight of a classifier
that makes many mistakes rapidly disappears.
We start with $\gamma=1.0$ and decrease its value
with the number of examples seen, to avoid weighing mistakes
of the initial hypotheses too heavily.\footnote{
  The exact form of the decreasing function is unimportant;
  we interpolate quadratically between $1.0$ and $0.67$
  as a decreasing function of the number of examples.}
The total activation returned by the cloud is then:
$$ \frac{\sum_j \gamma^{m_j} C_j}{\sum_j \gamma^{m_j}}, $$
where $C_j$ is the classification, either 1 or 0,
returned by the $j$th classifier in the cloud,
and the denominator is a normalization term.

The rationale for combining evidence from multiple classifiers is twofold.
First, when running a mistake-driven algorithm,
even when it is known to have good behavior asymptotically,
there is no guarantee that the current hypothesis, at any point in time,
is any better than the previous one.
It is common practice, therefore, to predict
using an average of the last several hypotheses,
weighting each hypothesis by, for example,
the length of its mistake-free stretch \cite{Littlestone95,CBFHW94}.
The second layer of \Winnow, i.e., the weighted-majority layer,
partly serves this function,
though it does so in an on-line fashion.

A second motivation for the weighted-majority layer comes from the desire to
have an algorithm that tunes its own parameters.
For the task of \spelling,
self-tuning is used to automatically accommodate differences
among confusion sets --- in particular, differences in the degree
to which the words in the confusion set have overlapping usages.
For \weather, for example, the words occur in essentially disjoint contexts;
thus, if a training example gives one word, but the classifier predicts
the other, it is almost surely wrong.
On the other hand, for \among, there are numerous contexts
in which both words are acceptable;
thus disagreement with the training example does not necessarily
mean the classifier is wrong.
Following a mistake, therefore, we want to demote the weights
by more in the former case than in the latter.
Updating weights with various demotion parameters in parallel
allows the algorithm to select by itself the best setting of
parameters for each confusion set.
In addition, using a weighted-majority layer strictly
increases the expressivity of the architecture.
It is plausible that in some cases, a linear separator
would be unable to achieve good prediction,
while the two-layer architecture would be able to do so.

\subsection{Discussion} \label{sec:winnowdiscuss}

Multiplicative learning algorithms have been studied extensively in the
learning theory community in recent years
\cite{Littlestone88,KivinenWa95}.  Winnow has been shown to learn
efficiently any linear threshold function \cite{Littlestone88}, with a
mistake bound that depends on the margin between positive and negative
examples. These are functions $f: \{0,1\}^{n} \ra \{0,1\}$ for which
there exist real weights $w_1, \ldots, w_n$ and a real threshold
$\theta$ such that $f(x_1,\ldots,x_n) = 1$ iff $\sum_{i=1}^{n} w_i x_i
\geq \theta$.  In particular, these functions include Boolean
disjunctions and conjunctions on $k \leq n$ variables and $r$-of-$k$
threshold functions ($1 \leq r \leq k \leq n$).  

The key feature of Winnow is that its mistake bound grows linearly
with the number of {\it relevant\/} attributes and only
logarithmically with the total number of attributes $n$. 
Using the sparse architecture, in which we do not keep all
the variables from the beginning, but rather add variables as
necessary, the number of mistakes made on disjunctions and
conjunctions is logarithmic in the size of the largest example seen and
linear in the number of relevant attributes; it is independent of the
total number of attributes in the domain \cite{Blum92}.

Winnow was analyzed in the presence of various kinds of noise, and in
cases where no linear threshold function can make perfect
classifications \cite{Littlestone91}.  It was proved, under some
assumptions on the type of noise, that Winnow still learns correctly,
while retaining its abovementioned dependence on the number of total
and relevant attributes.  
(See \namecite{KivinenWa95} for a thorough
analysis of multiplicative update algorithms versus additive update
algorithms, and exact bounds that depend on the sparsity of
the target function and the number of active features in the
examples.) 

The algorithm makes no independence or other
assumptions about the attributes, in contrast to Bayesian predictors
which are commonly used in statistical NLP.
This condition was recently investigated experimentally (on simulated
data) \cite{Littlestone95}.  It was shown that redundant attributes
dramatically affect a Bayesian predictor, while superfluous
independent attributes have a less dramatic effect, and only
when the number of attributes is very large (on the order of 10,000).
Winnow is a mistake-driven algorithm; that is, it updates its
hypothesis only when a mistake is made.  Intuitively, this makes the
algorithm more sensitive to the relationships among attributes ---
relationships that may go unnoticed by an algorithm that is based on
counts accumulated separately for each attribute.  This is crucial in
the analysis of the algorithm and has been shown to be crucial
empirically as well \cite{Littlestone95}.

One of the advantages of the multiplicative update algorithms is
their logarithmic dependence on the number of domain features. 
This property allows one to learn higher-than-linear discrimination
functions by increasing the dimensionality of the feature space. 
Instead of learning a discriminator in the original feature space,
one can generate new features, as conjunctions of original features,
and learn a linear separator in the new space, where it is more likely
to exist. Given the modest dependency of Winnow on the dimensionality,
it can be worthwhile to increase the dimensionality
so as to simplify the functional form of the resulting discriminator.
The work reported here can be regarded as following this path,
in that we define collocations as {\it patterns\/} of words
and part-of-speech tags, rather than restricting them
to tests of singleton elements.
This increases the dimensionality and adds redundancy among features,
but at the same time simplifies the functional form
of the discriminator, to the point that the classes
are almost linearly separable in the new space. 
A similar philosophy, albeit very different technically, is followed by 
the work on Support Vector Machines \cite{CortesVa95}.

Theoretical analysis has shown Winnow to be able to adapt quickly
to a changing target concept \cite{HerbsterWa95}.
We investigate this issue experimentally in Section~\ref{sec:across}.
A further feature of \Winnow\ is that it can prune
poorly-performing attributes, whose weight falls too low relative to
the highest weight of an attribute used by the classifier.
By pruning in this way, we can greatly reduce the number
of features that need to be retained in the representation.
It is important to observe, though, that there is a tension between
compacting the representation
by aggressively discarding features, and maintaining the ability
to adapt to a new test environment.
In this paper we focus on adaptation,
and do not study discarding techniques.
This tradeoff is currently under investigation.

\section{Experimental results} \label{sec:exp}

To understand the performance of \Winnow\ on the task of \spelling,
we start by comparing it with \Bayes\ using
the pruned set of features from the feature extractor,
which is what \Bayes\ normally uses.
This evaluates \Winnow\ purely as a method of combining evidence
from multiple features.
An important claimed strength of the Winnow-based approach, however,
is the ability to handle large numbers of features.  
We tested this by (essentially) disabling pruning,
resulting in tasks with over 10,000~features,
and seeing how \Winnow\ and \Bayes\ scale up.

The first experiment showed how \Winnow\ and \Bayes\ perform
relative to each other, but not to an outside reference.
To calibrate their performance,
we compared the two algorithms with other methods reported
in the literature, as well as a baseline method.

The success of \Winnow\ in the previous experiments
brought up the question of {\it why\/} it was able to
outperform \Bayes\ and the other methods.
We investigated this in an ablation study,
in which we stripped \Winnow\ down to a simple, non-learning algorithm,
and gave it an initial set of weights that
allowed it to emulate \Bayes's behavior exactly.
From there, we restored the missing aspects of \Winnow\ one at a time,
observing how much each contributed to
improving its performance above the Bayesian level.

The preceding experiments drew the training and test sets
from the same population, following the traditional PAC-learning assumption.
This assumption may be unrealistic for the task at hand, however,
where a system may encounter a target document
quite unlike those seen during training.
To check whether this was in fact a problem,
we tested the across-corpus performance of the methods.
We found it was indeed significantly worse
than within-corpus performance.
To address this problem,
we tried a strategy of combining learning on the training set
with unsupervised learning on the (noisy) test set.
We tested how well \Winnow\ and \Bayes\ were able to perform
on an unfamiliar test set using this strategy.

The sections below describe the experimental methodology,
and present the experiments, interleaved with discussion.

\subsection{Methodology} \label{sec:method}

In the experiments that follow,
the training and test sets were drawn from two corpora:
the 1-million-word Brown corpus \cite{brown-corpus}
and a 3/4-million-word corpus of articles
from The Wall Street Journal (WSJ) \cite{wsj-corpus}.
Note that no particular annotations are needed
on these corpora for the task of \spelling;
we simply assume that the texts
contain no context-sensitive spelling errors,
and thus the observed spellings may be taken
as a gold standard.

The algorithms were run on 21~confusion sets,
which were taken largely from the list of ``Words Commonly Confused''
in the back of the Random House dictionary \cite{random-house}.
The confusion sets were selected on the basis of being
frequently-occurring in Brown and WSJ, and include
mainly homophone confusions (e.g., \peace).
Several confusion sets not in Random House were added,
representing grammatical errors (e.g., \among)
and typographic errors (e.g., \maybe).

Results are reported as a percentage of correct classifications
on each confusion set, as well as an overall score, which
gives the percentage correct for all confusion sets pooled together.
When comparing scores, we tested for significance
using a McNemar test \cite{tgdtest} when possible;
when data on individual trials was not available
(the system comparison),
or the comparison was across different test sets
(the within/across study), we instead used a test for
the difference of two proportions \cite{fleiss}. All tests are reported for the 0.05 significance level.

\subsection{Pruned versus unpruned} \label{sec:prune}

The first step of the evaluation was to test \Winnow\ under
the same conditions that \Bayes\ normally runs under ---
i.e., using the pruned set of features from the feature extractor.
We used a random 80-20 split (by sentence) of Brown
for the training and test sets.
The results of running each algorithm on the 21~confusion sets
appear in the `Pruned' columns of Table~\ref{tab:prune}.
Although for a few confusion sets, one algorithm or the other does better,
overall \Winnow\ performs comparably to \Bayes.

\begin{table}[tbh]
\tabskip=3pt
\advance\tabskip by -\baselineskip
\begingroup\catcode`~=\active\def~{\kern\digitwidth}\caption{Pruned versus unpruned performance of \Bayes\ and \Winnow.
In the pruned condition, the algorithms use the pruned
set of features from the feature extractor;
in the unpruned condition, they use the full set
(excluding features occurring just once in the training set).
The algorithms were trained on 80\% of Brown and tested on the
other 20\%.
The first two columns give the number of features in the two conditions.
Bar graphs show the differences between adjacent columns,
with shading indicating significant differences
(using a McNemar test at the 0.05 level).
}
\def\tabvskip{& & &  & && & & && \\[\tabskip]}
\begin{tabular*}{\textwidth}{@{\extracolsep{\fill}}lcc
  @{\hspace*{23pt}} c \bgcol c
  @{\hspace*{23pt}} c \bgcol c}
\hline
Confusion set           & Pruned & Unpruned
  & \multicolumn{4}{c}{\hspace*{-10pt}Pruned}
  & \multicolumn{4}{c}{Unpruned} \\
                        & features   & features
  & \bglab{\Bayes} & \bgheader & \bglab{\Winnow}
  & \bglab{\Bayes} & \bgheader & \bglab{\Winnow} \\[3pt]
\savehline\tabvskip
accept, except          & ~~78 & ~~849 & 88.0 & \bar{0.2}& & 87.8 & 92.0 & &\bar{4.0} & 96.0 \\
affect, effect          & ~~36 & ~~842 & 98.0 & &\bar{2.0} & \hundred & 98.0 & &\bar{2.0} & \hundred \\
among, between          & ~145 & ~2706 & 75.3 & &\bar{0.5} & 75.8 & 78.0 & &\bar{8.0}[.5] & 86.0 \\
amount, number          & ~~68 & ~1618 & 74.8 & \bar{1.6}& & 73.2 & 80.5 & &\bar{5.7} & 86.2 \\
begin, being            & ~~84 & ~2219 & 95.2 & \bar{5.5}& & 89.7 & 95.2 & &\bar{2.7} & 97.9 \\
cite, sight, site       & ~~24 & ~~585 & 76.5 & \bar{11.8}& & 64.7 & 73.5 & &\bar{11.8} & 85.3 \\
country, county         & ~~40 & ~1213 & 88.7 & &\bar{1.3} & 90.0 & 91.9 & &\bar{3.3} & 95.2 \\
fewer, less             & ~~~6 & ~1613 & 96.0 & \bar{1.6}& & 94.4 & 92.0 & &\bar{1.3} & 93.3 \\
I, me                   & 1161 & 11625 & 97.8 & &\bar{0.4} & 98.2 & 98.3 & &\bar{0.2} & 98.5 \\
its, it's               & ~180 & ~4679 & 94.5 & &\bar{1.9} & 96.4 & 95.9 & &\bar{1.4} & 97.3 \\
lead, led               & ~~33 & ~~833 & 89.8 & \bar{2.3}& & 87.5 & 85.7 & &\bar{6.1} & 91.8 \\
maybe, may be           & ~~86 & ~1639 & 90.6 & \bar{6.2}[.5]& & 84.4 & 95.8 & &\bar{2.1} & 97.9 \\
passed, past            & ~141 & ~1279 & 90.5 & &          & 90.5 & 90.5 & &\bar{5.4} & 95.9 \\
peace, piece            & ~~67 & ~~992 & 74.0 & \bar{2.0}& & 72.0 & 92.0 & \bar{4.0}& & 88.0 \\
principal, principle    & ~~38 & ~~669 & 85.3 & \bar{0.5}& & 84.8 & 85.3 & &\bar{5.9} & 91.2 \\
quiet, quite            & ~~41 & ~1200 & 95.5 & \bar{0.1}& & 95.4 & 89.4 & &\bar{4.5} & 93.9 \\
raise, rise             & ~~24 & ~~621 & 79.5 & \bar{5.2}& & 74.3 & 87.2 & &\bar{2.5} & 89.7 \\
than, then              & ~857 & ~6813 & 93.6 & &\bar{3.3}[.5] & 96.9 & 93.4 & &\bar{2.3} & 95.7 \\
their, there, they're   & ~946 & ~9449 & 94.8 & &\bar{1.8} & 96.6 & 94.5 & &\bar{4.0}[.5] & 98.5 \\
weather, whether        & ~~61 & ~1226 & 93.4 & &\bar{5.0} & 98.4 & 98.4 & &\bar{1.6} & \hundred \\
your, you're            & ~103 & ~2738 & 90.4 & &\bar{3.2} & 93.6 & 90.9 & &\bar{6.4}[.5] & 97.3 \\[3pt]
\savehline\tabvskip
{\bf Overall}           &      &       & 93.0 & &\bar{0.7} & 93.7 & 93.8 & &\bar{2.6}[.5] & 96.4 \\[3pt]
\savehline
\end{tabular*}
\label{tab:prune}
\endgroup
\end{table}

The preceding comparison shows that \Winnow\ is
a credible method for this task,
but it does not test the claimed strength of Winnow ---
the ability to deal with large numbers of features.
To test this, we modified the feature extractor
to do only minimal pruning of features:
features were pruned only if they occurred exactly once in the training set
(such features are both extremely unlikely
to afford good generalizations, and extremely numerous).
The hope is that by considering the full set of features,
we will pick up many ``minor cases''
--- what \namecite{small-disjuncts} have called ``small disjuncts'' ---
that are normally filtered out by the pruning process.
The results are shown in the `Unpruned' columns of Table~\ref{tab:prune}.
While both algorithms do better in the unpruned condition,
\Winnow\ improves for almost every confusion set, sometimes markedly,
with the result that it outperforms \Bayes\ in the unpruned condition
for every confusion set except one.
The results below will all focus on
the behavior of the algorithms in the unpruned case.

\subsection{System comparison} \label{sec:syscom}

The previous section shows how \Winnow\ and \Bayes\ perform
relative to each other;
to evaluate them with respect to an external standard,
we compared them to other methods reported in the literature.
Two recent methods use some of the same test sets as we do,
and thus can readily be compared:
RuleS, a transformation-based learner \cite{manbri97};
and a method based on latent semantic analysis (LSA) \cite{jonmar97}.
We also compare to Baseline, the canonical straw man for this task,
which simply identifies the most common member of the confusion set
during training, and guesses it every time during testing.

\begin{table}[tbh]
\tabskip=3pt
\advance\tabskip by -\baselineskip
\begingroup\catcode`?=\active\def?{\kern\digitwidth}\caption{System comparison.
All algorithms were trained on 80\% of Brown and tested on the other 20\%;
all except LSA used the same 80-20 breakdown.
The version of RuleS is the one that uses the same feature set as we do.
\Bayes\ and \Winnow\ were run in the unpruned condition.
The first column gives the number of test cases.
Bar graphs show the differences between adjacent columns,
with shading indicating significant differences
(using a test for the difference of two proportions at the 0.05 level).
Ragged-ended bars indicate a difference of more than 15 percentage points.
The three `overall' lines pool the results over
different sets of confusion sets.
}
\def\tabvskip{& &   & && & && & && & && \\[\tabskip]}
\begin{tabular*}{\textwidth}  {@{\extracolsep{\fill}}l @{\hspace*{11pt}} c @{\hspace*{12pt}}
  c \bgcol c \bgcol c \bgcol c \bgcol c}
\hline
Confusion set           & Test & \bglab{Baseline} & \bgheader & \bglab{LSA} & \bgheader
                          & \bglab{RuleS} & \bgheader & \bglab{\Bayes} & \bgheader & \bglab{\Winnow} \\
                        & cases \\[3pt]
\savehline\tabvskip
accept, except          & ??50 & 70.0 & &\bar{12.3} & 82.3 & &\bar{5.7} & 88.0 & &\bar{4.0} & 92.0 & &\bar{4.0} & 96.0 \\
affect, effect          & ??49 & 91.8 & &\bar{2.5}  & 94.3 & &\bar{3.6} & 97.9 & &\bar{0.1} & 98.0 & &\bar{2.0} & \hundred \\
among, between          & ?186 & 71.5 & &\bar{9.3}  & 80.8 & \bar{7.7}& & 73.1 & &\bar{4.9} & 78.0 & &\bar{8.0} & 86.0 \\
amount, number          & ?123 & 71.5 & \bar{14.9}[.75]& & 56.6 & &\barlightrt& 78.0 & &\bar{2.5} & 80.5 & &\bar{5.7} & 86.2 \\
begin, being            & ?146 & 93.2 & &           & 93.2 & &\bar{2.1} & 95.3 & \bar{0.1}& & 95.2 & &\bar{2.7} & 97.9 \\
cite, sight, site       & ??34 & 64.7 & &\bar{13.4} & 78.1 & &          &      & &          & 73.5 & &\bar{11.8} & 85.3 \\
country, county         & ??62 & 91.9 & \bar{10.6}& & 81.3 & &\bar{13.9}[.75]& 95.2 & \bar{3.3}& & 91.9 & &\bar{3.3} & 95.2 \\
fewer, less             & ??75 & 90.7 & &           &      & &          &      & &          & 92.0 & &\bar{1.3} & 93.3 \\
I, me                   & 1225 & 83.0 & &           &      & &          &      & &          & 98.3 & &\bar{0.2} & 98.5 \\
its, it's               & ?366 & 91.3 & &\bar{1.5}  & 92.8 & &          &      & &          & 95.9 & &\bar{1.4} & 97.3 \\
lead, led               & ??49 & 46.9 & &\barlightrt  & 73.0 & &\barhollowrt& 89.8 & \bar{4.1}& & 85.7 & &\bar{6.1} & 91.8 \\
maybe, may be           & ??96 & 87.5 & &           &      & &          &      & &          & 95.8 & &\bar{2.1} & 97.9 \\
passed, past            & ??74 & 68.9 & &\bar{11.4} & 80.3 & &\bar{3.4} & 83.7 & &\bar{6.8} & 90.5 & &\bar{5.4} & 95.9 \\
peace, piece            & ??50 & 44.0 & &\barlightrt  & 83.9 & &\bar{6.1} & 90.0 & &\bar{2.0} & 92.0 & \bar{4.0}& & 88.0 \\
principal, principle    & ??34 & 58.8 & &\barlightrt  & 91.2 & \bar{3.0}& & 88.2 & \bar{2.9}& & 85.3 & &\bar{5.9} & 91.2 \\
quiet, quite            & ??66 & 83.3 & &\bar{7.5}  & 90.8 & &\bar{1.6} & 92.4 & \bar{3.0}& & 89.4 & &\bar{4.5} & 93.9 \\
raise, rise             & ??39 & 64.1 & &\barhollowrt & 80.6 & &\bar{4.0} & 84.6 & &\bar{2.6} & 87.2 & &\bar{2.5} & 89.7 \\
than, then              & ?514 & 63.4 & &\barlightrt  & 90.5 & &\bar{2.1} & 92.6 & &\bar{0.8} & 93.4 & &\bar{2.3} & 95.7 \\
their, there, they're   & ?850 & 56.8 & &\barlightrt  & 73.9 & &          &      & &          & 94.5 & &\bar{4.0}[.75] & 98.5 \\
weather, whether        & ??61 & 86.9 & \bar{1.8}&  & 85.1 & &\bar{8.3} & 93.4 & &\bar{5.0} & 98.4 & &\bar{1.6} & \hundred \\
your, you're            & ?187 & 89.3 & &\bar{2.1}  & 91.4 & &          &      & &          & 90.9 & &\bar{6.4}[.75] & 97.3 \\[3pt]
\savehline\tabvskip
{\bf Overall (14 sets)} & 1503 & 71.1 & &\bar{13.4}[.75] & 84.5 & &\bar{4.0}[.75] & 88.5 & &\bar{1.4} & 89.9 & &\bar{3.6}[.75] & 93.5 \\
{\bf Overall (18 sets)} & 2940 & 70.6 & &\bar{12.2}[.75] & 82.8 & &          &      & &          & 91.8 & &\bar{3.8}[.75] & 95.6 \\
{\bf Overall}           & 4336 & 74.8 & &           &      & &          &      & &          & 93.8 & &\bar{2.6}[.75] & 96.4 \\[3pt]
\savehline
\end{tabular*}
\endgroup
\label{tab:syscom}
\end{table}

The results appear in Table~\ref{tab:syscom}.
The scores for LSA, taken from \namecite{jonmar97},
are based on a different 80-20 breakdown of Brown than that
used by the other systems.
The scores for RuleS are for the version of that system
that uses the same feature set as we do.
The comparison shows \Winnow\ to have significantly higher
performance than the other systems.
Interestingly, however, Mangu and Brill
were able to improve RuleS's overall score from 88.5 to 93.3
(almost up to the level of \Winnow)
by making various clever enhancements to the feature set,
including using a tagger to assign a word its possible tags in context,
rather than merely using the word's complete tag set.
This suggests that \Winnow\ might get a similar boost
by adopting this enhanced set of features.

A note on the LSA system:
LSA has been reported to do its best for confusion sets
in which the words all have the same part of speech.
Since this does not hold for all of our confusion sets,
LSA's overall score was adversely affected.

\subsection{Ablation Study} \label{sec:ablation}

The previous sections demonstrate
the superiority of \Winnow\ over \Bayes\ for the task at hand,
but they do not explain {\it why\/} the Winnow-based algorithm does better.
At their core, \Winnow\ and \Bayes\ are both linear separators
\cite{Roth98};
is it that Winnow, with its multiplicative update rule,
is able to learn a better linear separator
than the one given by Bayesian probability theory?
Or is it that the non-Winnow enhancements of \Winnow,
particularly weighted-majority voting,
provide most of the leverage?
To address these questions, we ran an ablation study
to isolate the contributions of different aspects of \Winnow.

The study was based on the observation that the core computations of
Winnow and Bayesian classifiers are essentially isomorphic:
Winnow makes its decisions based on a weighted sum
of the observed features.
Bayesian classifiers make their decisions based not on a sum,
but on a product of likelihoods (and a prior probability) ---
but taking the logarithm of this functional form yields a linear function.
With this understanding,
we can start with the full \Bayes\ system;
strip it down to its Bayesian essence;
map this (by taking the log) to a simplified, non-learning version
of \Winnow\ that performs the identical computation;
and then add back the removed aspects of \Winnow,
one at a time, to understand how much each contributes
to eliminating the performance difference between
(the equivalent of) the Bayesian essence
and the full \Winnow\ system.

The experiment proceeds in a series of steps
that morph \Bayes\ into \Winnow:

\begin{description}

\item[\Bayes] The full \Bayes\ method, which includes
dependency resolution and interpolative smoothing.

\item[Simplified \Bayes] Like \Bayes, but without dependency resolution.
This means that all matching features,
even highly interdependent ones,
are used in the Bayesian calculation.
We do not strip \Bayes\ all the way down
to Naive Bayes, which would use MLE likelihoods,
because the performance would then be so poor as to be
unrepresentative of \Bayes\ --- and this would undermine
the experiment, which seeks to investigate how
\Winnow\ improves on \Bayes\ (not on a pale imitation thereof).

\item[Simplified \Winnow] This is a minimalist \Winnow, set up
to emulate the computation of Simplified \Bayes.
It has a 1-layer architecture (i.e., no Weighted Majority layer);
it uses a full network (not sparse);
it is initialized with Bayesian weights (to be explained momentarily);
and it does no learning (i.e., it does not update the Bayesian weights).
The Bayesian weights are simply the log of Simplified \Bayes's
likelihoods, plus a constant, to make them all non-negative
(as required by Winnow).
Occasionally, a likelihood will be 0.0, in which case we
smooth the log(likelihood) from $-\infty$ to a large negative
constant (we used $-500$).  In addition, we add a pseudo-feature
to Winnow's representation, which is active for every example,
and corresponds to the prior.  The weight for this feature is
the log of the prior.

\item[1-layer \Winnow] Like Simplified \Winnow, but adds learning.
This lets us see whether Winnow's multiplicative
update rule is able to improve on the Bayesian feature weights.
We ran learning for 5~cycles of the training set.

\item[2-layer \Winnow] Like 1-layer \Winnow, but adds
the weighted-majority voting layer to the architecture.

\item[(Bayesian) \Winnow] Replaces the full network of
2-layer \Winnow\ with a sparse network.
This yields the complete \Winnow\ algorithm, although its performance
is affected by the fact that it started with Bayesian,
not uniform weights.

\end{description}

\begin{table}[tb]
\renewcommand{\bglab}[1]{\makebox[35pt]{#1}}  \tabskip=3pt
\advance\tabskip by -\baselineskip
\caption{Ablation study.
Training was on 80\% of Brown and testing on the other 20\%.
The algorithms were run in the unpruned condition.
Bar graphs show the differences between adjacent columns,
with shading indicating significant differences
(using a McNemar test at the 0.05 level).
}
\def\tabvskip{& & && & && & && & && \\[\tabskip]}
\begin{tabular*}{\textwidth}  {@{\extracolsep{\fill}}l @{\hspace*{5pt}} c \bgcol c \bgcol c \bgcol c \bgcol c}
\hline
Confusion set & \bglab{\Bayes} & \bgheader & \bglab{Simplified}
    & \bgheader & \bglab{1-layer} & \bgheader & \bglab{2-layer}
    & \bgheader & \bglab{(Bayesian)} \\
              &        & \bgheader & \bglab{\Bayes}
    & \bgheader & \bglab{\Winnow} & \bgheader & \bglab{\Winnow}
    & \bgheader & \bglab{\Winnow} \\[3pt]
\savehline\tabvskip
accept, except         & 92.0  & &               &   92.0  & &\bar{2.0}      &   94.0  & \bar{4.0}&      &   90.0  & &\bar{6.0}      &   96.0 \\
affect, effect         & 98.0  & \bar{2.0}&      &   95.9  & &\bar{2.0}      &   98.0  & &               &   98.0  & &\bar{2.0}      &   \hundred \\
among, between         & 78.0  & &\bar{1.6}      &   79.6  & \bar{2.2}&      &   77.4  & &\bar{13.4}[.5] &   90.9  & \bar{1.6}&      &   89.2 \\
amount, number         & 80.5  & \bar{2.4}&      &   78.0  & &\bar{6.5}      &   84.6  & &\bar{4.1}      &   88.6  & \bar{3.3}&      &   85.4 \\
begin, being           & 95.2  & \bar{6.8}[.5]&  &   88.4  & &\bar{8.2}[.5]  &   96.6  & &\bar{2.1}      &   98.6  & &\bar{0.7}      &   99.3 \\
cite, sight, site      & 73.5  & &               &   73.5  & &\bar{5.9}      &   79.4  & \bar{2.9}&      &   76.5  & &\bar{11.8}     &   88.2 \\
country, county        & 91.9  & \bar{11.3}[.5]& &   80.6  & &\bar{11.3}     &   91.9  & &\bar{1.6}      &   93.5  & &\bar{3.2}      &   96.8 \\
fewer, less            & 92.0  & &\bar{2.7}      &   94.7  & \bar{1.3}&      &   93.3  & &\bar{2.7}      &   96.0  & &\bar{1.3}      &   97.3 \\
I, me                  & 98.3  & \bar{0.4}&      &   97.9  & &\bar{0.7}      &   98.6  & &\bar{0.5}      &   99.1  & &\bar{0.4}      &   99.5 \\
its, it's              & 95.9  & \bar{1.4}&      &   94.5  & &\bar{1.4}      &   95.9  & &\bar{2.5}[.5]  &   98.4  & \bar{0.5}&      &   97.8 \\
lead, led              & 85.7  & &\bar{6.1}      &   91.8  & \bar{4.1}&      &   87.8  & &               &   87.8  & &\bar{6.1}      &   93.9 \\
maybe, may be          & 95.8  & &\bar{1.0}      &   96.9  & \bar{1.0}&      &   95.8  & &\bar{3.1}      &   99.0  & &               &   99.0 \\
passed, past           & 90.5  & &\bar{2.7}      &   93.2  & \bar{1.4}&      &   91.9  & \bar{4.1}&      &   87.8  & &\bar{5.4}      &   93.2 \\
peace, piece           & 92.0  & \bar{8.0}&      &   84.0  & &\bar{4.0}      &   88.0  & \bar{4.0}&      &   84.0  & &\bar{4.0}      &   88.0 \\
principal, principle   & 85.3  & &               &   85.3  & \bar{2.9}&      &   82.4  & &\bar{2.9}      &   85.3  & &\bar{5.9}      &   91.2 \\
quiet, quite           & 89.4  & &\bar{7.6}      &   97.0  & \bar{4.5}&      &   92.4  & \bar{1.5}&      &   90.9  & &\bar{3.0}      &   93.9 \\
raise, rise            & 87.2  & \bar{7.7}&      &   79.5  & &\bar{2.6}      &   82.1  & &               &   82.1  & &\bar{7.7}      &   89.7 \\
than, then             & 93.4  & &\bar{2.3}[.5]  &   95.7  & \bar{0.4}&      &   95.3  & &\bar{1.8}      &   97.1  & \bar{0.4}&      &   96.7 \\
their, there, they're  & 94.5  & \bar{1.8}&      &   92.7  & &\bar{4.6}[.5]  &   97.3  & &\bar{0.8}      &   98.1  & &\bar{0.1}      &   98.2 \\
weather, whether       & 98.4  & \bar{1.6}&      &   96.7  & &\bar{1.6}      &   98.4  & &\bar{1.6}      &   \hundred  & &               &   \hundred \\
your, you're           & 90.9  & \bar{1.6}&      &   89.3  & &\bar{7.5}[.5]  &   96.8  & &\bar{1.1}      &   97.9  & &\bar{1.1}      &   98.9 \\[3pt]
\savehline\tabvskip
{\bf Overall}          & 93.8  & \bar{0.7}&      &   93.1  & &\bar{2.0}[.5]  &   95.1  & &\bar{1.5}[.5]  &   96.6  & &\bar{0.6}[.5]  &   97.2 \\[3pt]
\savehline
\end{tabular*}
\label{tab:ablation}
\end{table}

The ablation study used the same 80-20 breakdown of Brown
as in the previous section, and the unpruned feature set.
The results appear in Table~\ref{tab:ablation}.
Simplified \Winnow\ has been omitted from the table,
as its results are identical to those of Simplified \Bayes.

The primary finding is that all three measured aspects
of \Winnow\ contribute positively to its improvement over \Bayes;
the ranking, from strongest to weakest benefit, is
(1)~the update rule, (2)~the weighted-majority layer, and
(3)~sparse networks.
The large benefit afforded by the update rule
indicates that Winnow is able to improve considerably
on the Bayesian weights.
The likely reason that the Bayesian weights are not already optimal
is that the Bayesian assumptions --- conditional feature independence
and adequate data for estimating likelihoods ---
do not hold fully in practice.
The Winnow update rule can surmount these difficulties
by tuning the likelihoods via feedback to fit
whatever situation holds in the (imperfect) world.
The feedback is obtained from the same training set
that is used to set the Bayesian likelihoods. 
Incidentally, it is interesting to note that the use of a sparse network
improves accuracy fairly consistently across confusion sets.
The reason it improves accuracy is that,
by omitting links for features that never co-occurred
with a given target word during training,
it effectively sets the weight of such features to 0.0,
which is apparently better for accuracy than
setting the weight to the log of the Bayesian likelihood
(which, in this case, is some {\it smoothed\/} version
of the 0.0 MLE probability).

A second observation concerns the performance of
\Winnow\ when starting with the Bayesian weights:
its overall score was 97.2\%, as compared with 96.4\%
for \Winnow\ when starting with uniform weights
(see Table~\ref{tab:syscom}).
This suggests that the performance of Winnow can be improved
by moving to a hybrid approach in which Bayes is used
to initialize the network weights.
This hybrid approach is also an improvement over Bayes:
in the present experiment, the pure Bayesian approach scored 93.1\%,
whereas when updates were performed on the Bayesian weights,
the score increased to 95.1\%.

A final observation from this experiment is that,
while it was intended primarily as an ablation study of \Winnow,
it also serves as a mini-ablation study of \Bayes.
The difference between the \Bayes\ and Simplified \Bayes\ columns
measures the contribution of dependency resolution.
It turns out to be almost negligible, which, at first glance,
seems surprising, considering the level of redundancy
in the (unpruned) set of features being used.
For instance, if the features include the collocation
``{\it a\/}~\cmark~{\it treaty\/}'',
they will also include collocations such as
``\tag{det}~\cmark~{\it treaty\/}'',
``{\it a\/}~\cmark~\tag{noun}$_{\rm sing}$'', and so on.
Nevertheless, there are two reasons that dependency resolution
is of little benefit.
First, the features are generated {\it systematically\/} by
the feature extractor,
and thus tend to overcount evidence equally for all words.
Second, Naive Bayes is known to be less sensitive to
the conditional independence assumption when we only
ask it to predict the most probable class (as we do here),
as opposed to asking it to predict the exact probabilities for all classes
\cite{DudaHa73,dompaz97}.
The contribution of interpolative smoothing --- the other
enhancement of \Bayes\ over Naive Bayes ---
was not addressed in Table~\ref{tab:ablation}.
However, we investigated this briefly
by comparing the performance of \Bayes\ with interpolative smoothing
to its performance with MLE likelihoods (the naive method),
as well as a number of alternative smoothing methods.
Table~\ref{tab:smoothing} gives the overall scores.
While the overall score for \Bayes\ with interpolative smoothing was 93.8\%,
it dropped to 85.8\% with MLE likelihoods,
and was also lower when alternative smoothing methods were tried.
This shows that while dependency resolution does not
improve \Bayes\ much over Naive Bayes,
interpolative smoothing does have a sizable benefit.

\begin{table}[tb]
\caption{Overall score for \Bayes\ using different smoothing methods.
The last method, interpolative smoothing, is the one presented here.
Training was on 80\% of Brown and testing on the other 20\%.
When using MLE likelihoods, we broke ties
by choosing the word with the largest prior
(ties arose when all words had probability 0.0).
For Katz smoothing, we used absolute discounting \cite{nek94},
as Good-Turing discounting resulted in invalid discounts for our task.
For Kneser-Ney smoothing, we used absolute discounting and
the backoff distribution based on the ``marginal constraint''.
For interpolation with a fixed $\lambda$, Katz, and Kneser-Ney,
we set the necessary parameters separately for each word $W_i$
using deleted estimation.
}
\begin{tabular*}{\textwidth}{l@{\hspace{.975in}}l@{\hspace{.975in}}c}
\hline
Smoothing method & Reference & {\bf Overall} \\
\hline
MLE likelihoods & & 85.8 \\
Interpolation with a fixed $\lambda$ & \namecite{nek94} & 89.8 \\
Laplace-$m$  & \namecite{kobeso97}     & 90.9  \\
No-matches-0.01 & \namecite{kobeso97} & 91.0 \\
Katz smoothing & \namecite{katz87}    & 91.6 \\
Kneser-Ney smoothing & \namecite{kneney95} & 93.4 \\
Interpolative smoothing & Section~\ref{sec:bayes} & 93.8 \\
\hline
\end{tabular*}
\label{tab:smoothing}
\end{table}

\subsection{Across-corpus performance} \label{sec:across}

The preceding experiments
assumed that the training set will be representative of the test set.
For \spelling, however, this assumption may be overly strong;
this is because word usage patterns vary widely
from one author to another, or even one document to another.
For instance, an algorithm may have been trained on one corpus
to discriminate between {\it desert\/} and {\it dessert},
but when tested on an article about the Persian Gulf War,
will be unable to detect the misspelling of {\it desert\/} in
{\it Operation Dessert Storm}.
To check whether this is in fact a problem,
we tested the across-corpus performance of the algorithms:
we again trained on 80\% of Brown,
but tested on a randomly-chosen 40\% of the sentences of WSJ.
The results appear in Table~\ref{tab:degradation}.
Both algorithms were found to degrade significantly.
At first glance, the magnitude of the degradation seems small ---
from 93.8\% to 91.2\% for the overall score of \Bayes,
and 96.4\% to 95.2\% for \Winnow.
However, when viewed as an increase in the error rate,
it is actually fairly serious:
for \Bayes, the error rate goes from 6.2\% to 8.8\% (a 42\% increase),
and for \Winnow, from 3.6\% to 4.8\% (a 33\% increase).
In this section, we present a strategy for dealing with
the problem of unfamiliar test sets,
and we evaluate its effectiveness when used by \Winnow\ and \Bayes.

\begin{table}[tbh]
\tabskip=3pt
\advance\tabskip by -\baselineskip
\begingroup\catcode`?=\active\def?{\kern\digitwidth}\caption{Within- versus across-corpus performance of \Bayes\ and \Winnow.
Training was on 80\% of Brown in both cases.
Testing for the within-corpus case was on 20\% of Brown;
for the across-corpus case, it was on 40\% of WSJ.
The algorithms were run in the unpruned condition.
Bar graphs show the differences between adjacent columns,
with shading indicating significant differences
(using a test for the difference of two proportions at the 0.05 level).
Ragged-ended bars indicate a difference of more than 15 percentage points.
}
\def\tabvskip{& & &  & && & & && \\[\tabskip]}
\begin{tabular*}{\textwidth}{@{\extracolsep{\fill}}l
  @{\hspace*{0pt}}c@{\hspace*{6pt}}c
  @{\hspace*{25pt}} c \bgcol c
  @{\hspace*{25pt}} c \bgcol c}
\hline
Confusion set           & Test cases & Test cases
  & \multicolumn{4}{c}{\hspace*{-16pt}\Bayes}
  & \multicolumn{4}{c}{\hspace*{5pt}\Winnow} \\
                        & Within     & Across 
  & \bglab{Within} & \bgheader & \bglab{Across}
  & \bglab{Within} & \bgheader & \bglab{Across} \\[3pt]
\savehline\tabvskip
accept, except       & ??50 & ??30 & 92.0  & \bar{12.0}&     &    80.0 &    96.0  & \bar{2.7}&      &    93.3 \\
affect, effect       & ??49 & ??66 & 98.0  & \bar{10.1}&     &    87.9 &   \hundred  & \bar{4.5}&      &    95.5 \\
among, between       & ?186 & ?256 & 78.0  & &\bar{1.3}      &    79.3 &    86.0  & &\bar{1.1}      &    87.1 \\
amount, number       & ?123 & ?167 & 80.5  & \bar{11.0}[.75]& &   69.5 &    86.2  & \bar{12.5}[.75]& &    73.7 \\
begin, being         & ?146 & ?174 & 95.2  & \bar{6.1}&      &   89.1 &    97.9  & &\bar{0.9}      &    98.9 \\
cite, sight, site    & ??34 & ??18 & 73.5  & \barhollowleft&  &   50.0 &    85.3  & \barlightleft& &    55.6 \\
country, county      & ??62 & ??71 & 91.9  & &\bar{2.5}      &    94.4 &    95.2  & &\bar{0.6}      &    95.8 \\
fewer, less          & ??75 & ?148 & 92.0  & &\bar{2.6}      &    94.6 &    97.3  & \bar{0.0}&      &    97.3 \\
I, me                & 1225 & ?328 & 98.3  & \bar{0.4}&      &    97.9 &    97.9  & \bar{5.4}&      &    92.5 \\
its, it's            & ?366 & 1277 & 95.9  & \bar{0.4}&      &    95.5 &    93.3  & &\bar{2.6}      &    95.9 \\
lead, led            & ??49 & ??69 & 85.7  & \bar{6.0}&      &    79.7 &    98.5  & \bar{0.1}&      &    98.5 \\
maybe, may be        & ??96 & ??67 & 95.8  & \bar{3.3}&      &    92.5 &    91.8  & \bar{2.0}&      &    89.9 \\
passed, past         & ??74 & ?148 & 90.5  & &\bar{5.4}      &    95.9 &    95.9  & &\bar{2.0}      &    98.0 \\
peace, piece         & ??50 & ??19 & 92.0  & \bar{13.1}&     &    78.9 &    88.0  & \bar{14.3}&     &    73.7 \\
principal, principle & ??34 & ??30 & 85.3  & \barhollowleft& &    70.0 &    91.2  & \bar{4.5}&      &    86.7 \\
quiet, quite         & ??66 & ??20 & 89.4  & \barlightleft& &    65.0 &    93.9  & \barlightleft& &    75.0 \\
raise, rise          & ??39 & ?118 & 87.2  & \barhollowleft& &    72.0 &    89.7  & \bar{7.5}&      &    82.2 \\
than, then           & ?514 & ?637 & 93.4  & &\bar{3.1}[.75]  &    96.5 &    95.7  & &\bar{2.7}[.75]  &    98.4 \\
their, there, they're& ?850 & ?748 & 94.5  & \bar{2.8}[.75]&  &    91.7 &    98.5  & \bar{0.3}&      &    98.1 \\
weather, whether     & ??61 & ??95 & 98.4  & \bar{3.7}&      &    94.7 &   \hundred  & \bar{3.2}&      &    96.8 \\
your, you're         & ?187 & ??74 & 90.9  & \bar{5.8}&      &    85.1 &    97.3  & \bar{1.4}&      &    95.9 \\[3pt]
\savehline\tabvskip
{\bf Overall}        & 4336 & 4560 & 93.8  & \bar{2.6}[.75]&  &    91.2 &    96.4  & \bar{1.3}[.75]&  &    95.2 \\[3pt]
\savehline
\end{tabular*}
\endgroup
\label{tab:degradation}
\end{table}

The strategy is based on the observation that the test document,
though imperfect, still provides a valuable source of information
about its own word usages.  Returning to the Desert Storm example,
suppose the system is asked to correct an article
containing 17~instances of the phrase {\it Operation Desert Storm},
and 1~instance of the (erroneous) phrase {\it Operation Dessert Storm}.
If we treat the {\it test\/} corpus as a training document,
we will then start by running the feature extractor,
which will generate (among others) the collocation:
\beginsentences
\sitem {\it Operation\/} \cmark~{\it Storm}.        \label{sent:storm}
\endsentences
The algorithm, whether \Bayes\ or \Winnow,
should then be able to learn,
during its training on the test (qua training) corpus,
that feature~\ref{sent:storm} typically co-occurs with
{\it desert}, and is thus evidence in favor of that word.
The algorithm can then use this feature to fix the one erroneous
spelling of the phrase in the test set.

It is important to recognize that the system is not ``cheating''
by looking at the test set; it would be cheating
if it were given an answer key along with the test set.
What the system is really doing is enforcing consistency
across the test set.  It can detect sporadic errors,
but not systematic ones (such as writing {\it Operation Dessert Storm\/}
every time).  However, it should be possible
to pick up at least some systematic errors by also doing
regular supervised learning on a training set.

This leads to a strategy, which we call {\it sup/unsup},
of combining supervised learning on the training set
with unsupervised learning on the (noisy) test set.
The learning on the training set is {\it supervised\/}
because a benevolent teacher ensures that all spellings
are correct (we establish this simply by assumption).
The learning on the test set is {\it unsupervised\/}
because no teacher tells the system
whether the spellings it observes are right or wrong.

We ran both \Winnow\ and \Bayes\ with sup/unsup to see
the effect on their across-corpus performance.
We first needed a test corpus containing errors;
we generated one by corrupting a correct corpus.
We varied the amount of corruption from 0\% to 20\%,
where $p$\% corruption means we altered a randomly-chosen $p$\%
of the occurrences of the confusion set
to be a different word in the confusion set.

The sup/unsup strategy calls for training on both
a training corpus and a corrupted test corpus,
and testing on the uncorrupted test corpus.
For purposes of this experiment, however,
we split the test corpus into two parts
to avoid any confusion about training and testing on the same data.
We trained on 80\% of Brown plus a corrupted version
of 60\% of WSJ; and we tested on the uncorrupted version of
the other 40\% of WSJ.

\begin{table}[tb]
\tabskip=3pt
\advance\tabskip by -\baselineskip
\begingroup\catcode`?=\active\def?{\kern\digitwidth}\caption{Across-corpus performance of \Bayes\ and \Winnow\ using
the sup/unsup strategy.
Performance is compared with doing supervised learning only.
Training in the sup/unsup case
is on 80\% of Brown plus 60\% of WSJ (5\% corrupted);
in the supervised case, it is on 80\% of Brown only.
Testing in all cases is on 40\% of WSJ.
The algorithms were run in the unpruned condition.
Bar graphs show the differences between adjacent columns,
with shading indicating significant differences
(using a McNemar test at the 0.05 level).
Ragged-ended bars indicate a difference of more than 15 percentage points.
}
\def\tabvskip{& &  & && & & && \\[\tabskip]}
\begin{tabular*}{\textwidth}{@{\extracolsep{\fill}}l
  @{\hspace*{10pt}}c
  @{\hspace*{35pt}} c \bgcol c
  @{\hspace*{35pt}} c \bgcol c}
\hline
Confusion set           & Test cases
  & \multicolumn{4}{c}{\hspace*{-30pt}\Bayes}
  & \multicolumn{4}{c}{\hspace*{5pt}\Winnow} \\
                        & 
  & \bglab{Sup only} & \bgheader & \bglab{Sup/unsup}
  & \bglab{Sup only} & \bgheader & \bglab{Sup/unsup} \\[3pt]
\savehline\tabvskip
accept, except          & ??30   & 80.0 & &\bar{6.7} & 86.7  & 93.3 & \bar{6.6}&  & 86.7 \\
affect, effect          & ??66   & 87.9 & &\bar{3.0} & 90.9  & 95.5 & \bar{1.4}&  & 93.9 \\
among, between          & ?256   & 79.3 & &\bar{1.9} & 81.2  & 87.1 & &\bar{3.5}  & 90.6 \\
amount, number          & ?167   & 69.5 & &\bar{8.9}[.5] & 78.4  & 73.7 & &\bar{13.7}[.5] & 87.4 \\
begin, being            & ?174   & 89.1 & &\bar{5.2}[.5] & 94.3  & 98.9 & &\bar{0.5} & 99.4 \\
cite, sight, site       & ??18   & 50.0 & &\barhollowrt & 66.7  & 55.6 & &\barhollowrt & 72.2 \\
country, county         & ??71   & 94.4 & &\bar{1.4} & 95.8  & 95.8 & &\bar{1.4}  & 97.2 \\
fewer, less             & ?148   & 94.6 & \bar{1.4}& & 93.2  & 95.9 & &\bar{2.1}  & 98.0 \\
I, me                   & ?328   & 97.9 & &\bar{0.6} & 98.5  & 98.5 & &\bar{0.6}  & 99.1 \\
its, it's               & 1277   & 95.5 & &\bar{0.1} & 95.6  & 97.3 & &\bar{0.5}  & 97.8 \\
lead, led               & ??69   & 79.7 & \bar{4.3}& & 75.4  & 89.9 & \bar{1.5}&  & 88.4 \\
maybe, may be           & ??67   & 92.5 & \bar{1.5}& & 91.0  & 92.5 & &\bar{4.5}  & 97.0 \\
passed, past            & ?148   & 95.9 & &\bar{0.7} & 96.6  & 98.0 & &           & 98.0 \\
peace, piece            & ??19   & 78.9 & &\bar{5.3} & 84.2  & 73.7 & &\barhollowrt & 89.5 \\
principal, principle    & ??30   & 70.0 & &\bar{6.7} & 76.7  & 86.7 & &\bar{3.3}   & 90.0 \\
quiet, quite            & ??20   & 65.0 & &\bar{10.0} & 75.0  & 75.0 & &\bar{15.0} & 90.0 \\
raise, rise             & ?118   & 72.0 & &\bardarkrt & 87.3  & 82.2 & &\bar{7.6}  & 89.8 \\
than, then              & ?637   & 96.5 & \bar{0.3}& & 96.2  & 98.4 & \bar{0.1}&   & 98.3 \\
their, there, they're   & ?748   & 91.7 & \bar{0.9}& & 90.8  & 98.1 & &\bar{0.4}   & 98.5 \\
weather, whether        & ??95   & 94.7 & &\bar{1.1} & 95.8  & 96.8 & &            & 96.8 \\
your, you're            & ??74   & 85.1 & &\bar{2.7} & 87.8  & 95.9 & &\bar{1.4}   & 97.3 \\[3pt]
\savehline\tabvskip
{\bf Overall}           & 4560   & 91.2 & &\bar{1.2}[.5] & 92.4  & 95.2 & &\bar{1.4}[.5] & 96.6 \\[3pt]
\savehline
\end{tabular*}
\endgroup
\label{tab:supunsup}
\end{table}

The results for the 5\% level of corruption are shown
in Table~\ref{tab:supunsup};
this level of corruption corresponds to 
typical typing error rates.\footnote{
  \namecite{mdm91}, for example,
  consider error rates from 0.01\% to 10\% for the same task.}
The table compares across-corpus performance of each algorithm
with and without the additional boost of unsupervised learning
on part of the test corpus.
Both \Bayes\ and \Winnow\ benefit from the unsupervised learning
by about the same amount;
the difference is that \Winnow\ suffered considerably less
than \Bayes\ when moving from the within- to the across-corpus condition.
As a result, \Winnow, unlike \Bayes,
is actually able to recover to its within-corpus performance level,
when using the sup/unsup strategy in the across-corpus condition.

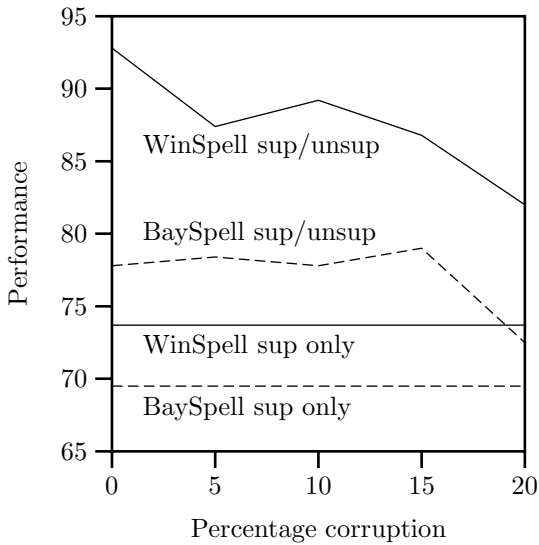
\begin{figure}[t]
\leftskip=.6in %\input{raise-graph.tex}
% GNUPLOT: LaTeX picture with Postscript
\setlength{\unitlength}{0.1bp}
\begin{picture}(2339,1943)(0,0)
\put(717,1400){\makebox(0,0)[l]{\Winnow\ sup/unsup}}
\put(717,648){\makebox(0,0)[l]{\Winnow\ sup only}}
\put(717,1072){\makebox(0,0)[l]{\Bayes\ sup/unsup}}
\put(717,415){\makebox(0,0)[l]{\Bayes\ sup only}}
\put(1378,-49){\makebox(0,0){Percentage corruption}}
\put(280,1071){%
%
\makebox(0,0)[b]{\shortstack{Performance}}%
%
}
\put(2156,151){\makebox(0,0){\raisebox{-12pt}{$20$}}}
\put(1767,151){\makebox(0,0){\raisebox{-12pt}{$15$}}}
\put(1378,151){\makebox(0,0){\raisebox{-12pt}{$10$}}}
\put(989,151){\makebox(0,0){\raisebox{-12pt}{$5$}}}
\put(600,151){\makebox(0,0){\raisebox{-12pt}{$0$}}}
\put(540,1892){\makebox(0,0)[r]{$95$ }}
\put(540,1619){\makebox(0,0)[r]{$90$ }}
\put(540,1345){\makebox(0,0)[r]{$85$ }}
\put(540,1072){\makebox(0,0)[r]{$80$ }}
\put(540,798){\makebox(0,0)[r]{$75$ }}
\put(540,525){\makebox(0,0)[r]{$70$ }}
\put(540,251){\makebox(0,0)[r]{$65$ }}
\end{picture}
\vspace*{0.2in}
\caption{Across-corpus performance of \Bayes\ (dotted lines)
and \Winnow\ (solid lines)
with the sup/unsup strategy
and with supervised learning only.
The curves show performance as a function
of the percentage corruption of the test set.
Training in the sup/unsup case is on 80\% of Brown,
plus 60\% of WSJ (corrupted); for the supervised-only case,
it is on 80\% of Brown only.
Testing in both cases is on 40\% of WSJ.
The algorithms were run for the confusion set \amount\ in the
unpruned condition.}
\label{fig:supunsup}
\end{figure}

It should be borne in mind that the results in Table~\ref{tab:supunsup}
depend on two factors.  The first is the size of the test set:
the larger the test set, the more information it can provide
during unsupervised learning.
The second factor is the percentage corruption of the test set.
Figure~\ref{fig:supunsup} shows performance
as a function of percentage corruption for a representative
confusion set, \amount.  As one would expect, the improvement
from unsupervised learning tends to decrease
as the percentage corruption increases.
For \Bayes's performance on \amount,
20\% corruption is almost enough
to negate the benefit of unsupervised learning.

\section{Conclusion} \label{sec:concl}

While theoretical analyses of the Winnow family of algorithms
have predicted an excellent ability to deal with large
numbers of features and to adapt to new trends
not seen during training,
these properties have remained largely undemonstrated.
In the work reported here, we have presented an architecture
based on Winnow and Weighted Majority,
and applied it to a real-world task, \spelling,
that has a potentially huge number of features
(over 10,000 in some of our experiments).
We showed that our algorithm, \Winnow, performs significantly better
than other methods tested on this task with a comparable feature set.
When comparing \Winnow\ to \Bayes, a Bayesian statistics-based algorithm
representing the state of the art for this task,
we found that \Winnow's mistake-driven update rule,
its use of weighted-majority voting,
and its sparse architecture
all contributed significantly to its superior performance.

\Winnow\ was found to exhibit two striking advantages over
the Bayesian approach.
First, \Winnow\ was substantially more accurate
than \Bayes\ when running with full (unpruned) feature sets,
outscoring \Bayes\ on 20 out of 21 confusion sets,
and achieving an overall score of over 96\%.
Second, \Winnow\ was better than \Bayes\ at adapting
to an unfamiliar test corpus, when using a strategy we presented
that combines supervised learning on the training set
with unsupervised learning on the test set.

This work represents an application of techniques developed
within the theoretical learning community in recent years,
and touches upon some of the important issues still under active research.
First, it demonstrates the ability of a Winnow-based algorithm
to successfully utilize the strategy of expanding the space of features
in order to simplify the functional form of the discriminator;
this was done in generating collocations as {\it patterns\/}
of words and part-of-speech tags.
The use of this strategy in Winnow shares much the same philosophy
--- if none of the technical underpinnings ---
as Support Vector Machines \cite{CortesVa95}.
Second, the two-layer architecture used here is related to various voting
and boosting techniques studied in recent years in the learning
community \cite{boosting95,bagging,LittlestoneWa94}. 
The goal is to learn to combine simple learners in
a way that improves the overall performance of the system.
The focus in the work reported here is on doing this learning
in an on-line fashion.

There are many issues still to investigate in order to develop
a complete understanding of the use of multiplicative update
algorithms in real-world tasks.
One of the important issues this work raises is
the need to understand and improve the ability of algorithms
to adapt to unfamiliar test sets.
This is clearly a crucial issue for algorithms to be used in real systems.
A related issue is that of the size and comprehensibility
of the output representation.
\namecite{manbri97}, using a similar set of features
to the one used here, demonstrate that massive feature pruning
can lead to highly compact classifiers, with surprisingly little
loss of accuracy.
There is a clear tension, however, between achieving a compact
representation and retaining the ability to adapt to unfamiliar test sets.
Further analysis of this tradeoff is under investigation.

The Winnow-based approach presented in this paper is being developed as
part of a research program in which we are trying to understand
how networks of simple and slow neuron-like elements can encode a
large body of knowledge and perform a wide range of interesting
inferences almost instantaneously.
We investigate this question in the context of learning
knowledge representations that support language understanding tasks.
In light of the encouraging results presented here for \spelling,
as well as other recent results \cite{DaganKaRo97,LiereTa97,RothZe98},
we are now extending the approach to other tasks.

\acknowledgements

We would like to thank Neal Lesh, Grace Ngai, Stan Chen, the reviewers,
and the editors
for helpful comments on the paper.
The second author's research was supported by
the Feldman Foundation and a grant from
the Israeli Ministry of Science and the Arts;
it was done partly while at Harvard University
supported by NSF grant CCR-92-00884
and DARPA contract AFOSR-F4962-92-J-0466.

\end{article}
\end{document}